\documentclass[letterpaper, 10 pt, conference]{ieeeconf}
\usepackage{amsmath,amsfonts}
\usepackage{array}
\usepackage{enumerate}
\usepackage[utf8]{inputenc}
\usepackage[T1]{fontenc}
\usepackage{amsfonts}
\usepackage{amssymb}
\IEEEoverridecommandlockouts 
\usepackage{algorithm}  
\usepackage{algpseudocode}  
\usepackage{amsmath}  
\usepackage{breqn}
\usepackage[caption=false]{subfig}
\usepackage{textcomp}
\usepackage{stfloats}
\usepackage{url}
\usepackage{verbatim}
\usepackage{graphicx}
\usepackage{cite}

\usepackage{hyperref}

\title{\LARGE \bf
Feedback-efficient Active Preference Learning for \\ Socially Aware Robot Navigation
}

\author{Ruiqi Wang$^{1}$, Weizheng Wang$^{1,2}$, and Byung-Cheol Min$^{1}$ 
\thanks{$^{1}$SMART Laboratory, Department of Computer and Information Technology, Purdue University, West Lafayette, IN, USA. {\tt\small{[wang5357,minb]@purdue.edu}.}}
\thanks{$^{2}$College of Mechanical and Electrical Engineering, Beijing University of Chemical Technology, Beijing (BUCT), China. \tt\small{wz.w.robot@gmail.com}.}}

\begin{document}

\maketitle

\begin{abstract}
Socially aware robot navigation, where a robot is required to optimize its trajectory to maintain comfortable and compliant spatial interactions with humans in addition to reaching its goal without collisions, is a fundamental yet challenging task in the context of human-robot interaction. While existing learning-based methods have achieved better performance than the preceding model-based ones, they still have drawbacks: reinforcement learning depends on the handcrafted reward that is unlikely to effectively quantify broad social compliance, and can lead to reward exploitation problems; meanwhile, inverse reinforcement learning suffers from the need for expensive human demonstrations. In this paper, we propose a feedback-efficient active preference learning approach, FAPL, that distills human comfort and expectation into a reward model to guide the robot agent to explore latent aspects of social compliance. We further introduce hybrid experience learning to improve the efficiency of human feedback and samples, and evaluate benefits of robot behaviors learned from FAPL through extensive simulation experiments and a user study (N=10) employing a physical robot to navigate with human subjects in real-world scenarios. Source code and experiment videos for this work are available at: \url{https://sites.google.com/view/san-fapl}.
\end{abstract}

\section{Introduction}

\begin{figure}[!t]
\centering
\includegraphics[width=0.8\columnwidth]{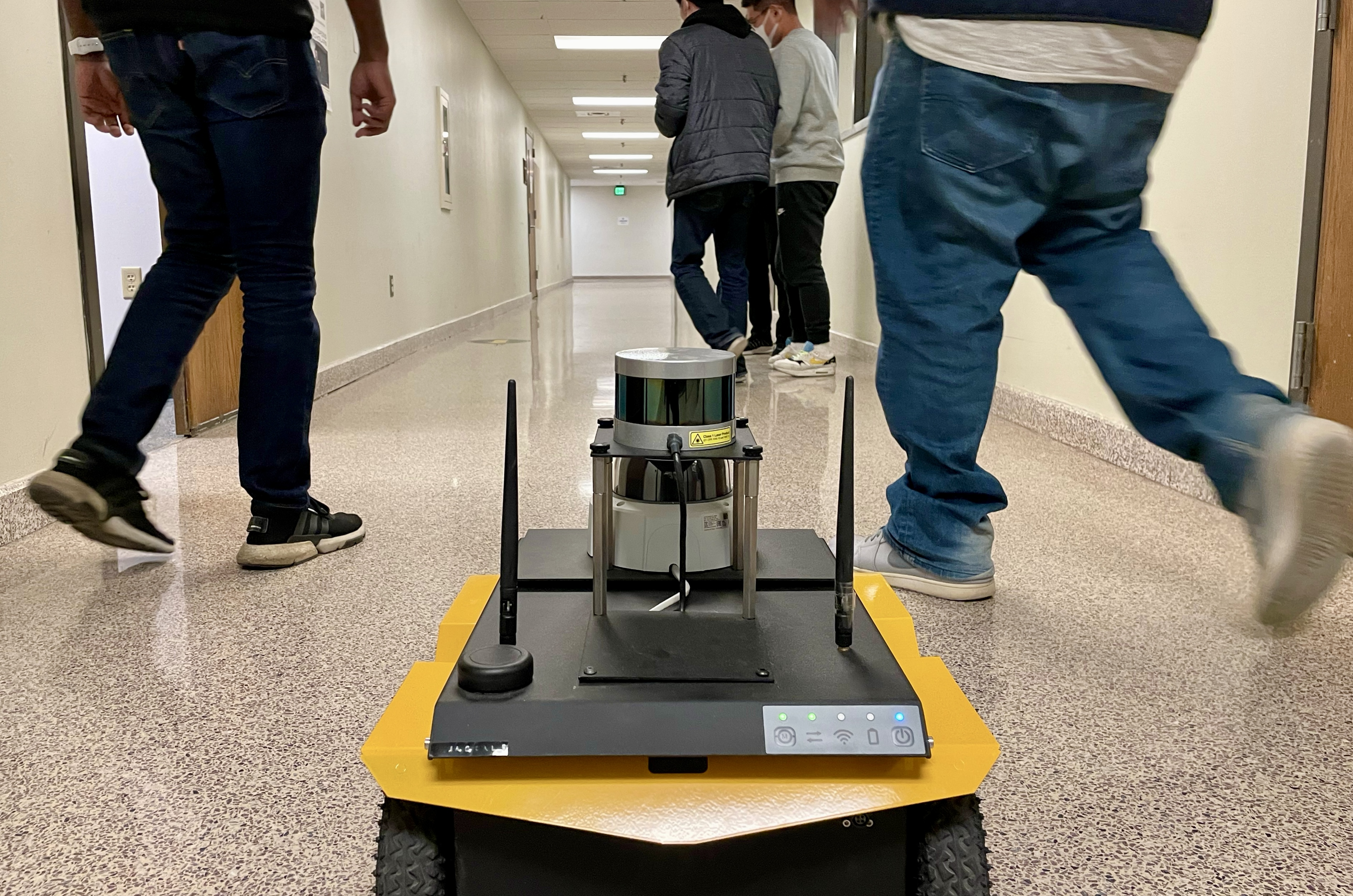}
\vspace{-7pt}
\caption{An example of real-world socially aware navigation with a mobile robot in a human-filled environment.}
\vspace{5pt}
\label{fig:mobile_robot}
\end{figure}

Advances in artificial intelligence-embedded robotics are increasingly enabling robots to work in environments that necessitate human-robot interaction (HRI). Delivery robots around university campuses, guide robots in shopping malls, elder care robots at nursing homes, and other such applications all require robots to perform socially aware navigation in human-rich environments, wherein the robots must not only consider how to complete navigation tasks successfully but also recognize and follow social etiquette to sustain comfortable spatial interaction with humans \cite{rios2015proxemics,kruse2013human}. For example, when navigating in a human-filled environment as depicted in Fig. \ref{fig:mobile_robot}, beyond simply reaching the final goal without collisions, the robot must maintain an acceptable distance from other pedestrians and adjust its movements to generate a comfortable interaction experience for humans. 

\setlength{\textfloatsep}{0pt}

Existing research in the field of socially aware navigation can be divided into two main categories: model-based and learning-based approaches. Model-based methods \cite{van2011reciprocal,vemula2018social,ferrer2014behavior}  aim to model the social conventions and dynamics of crowds to serve as additional parameters for traditional multi-agent collision-free robot navigation algorithms. However, it is quite difficult to emulate with one single model the precise rules followed by all pedestrians~\cite{rios2015proxemics}, not to mention that oscillatory trajectories can be produced \cite{chen2017socially}.


As an alternative, learning-based approaches have achieved better performance \cite{rios2015proxemics}. Within this category, there are two main types: reinforcement learning (RL) \cite{chen2017socially,chen2019crowd,chen2020relational} and inverse reinforcement learning (IRL) \cite{kretzschmar2016socially,okal2016learning,kim2016socially}. A primary shortcoming of RL methods is that engineering a meticulous handcrafted reward which emulates non-quantifiable social compliance is a non-trivial endeavor. Another problem stemming form handcrafted rewards is reward exploitation, that is, robots learn to achieve high rewards via some undesired and unnatural action that impairs human comfort. On the other hand, IRL methods, where a policy or reward is learned from human demonstrations, can avoid reward engineering and exploitation and allow experts to introduce human insights and comfort into robot policy. However, obtaining sufficient and accurate demonstrations is expensive, and careful feature engineering is required to achieve sound performance \cite{tsai2020generative}.

\begin{figure*}[!t]
\centering
\includegraphics[width=0.94\linewidth]{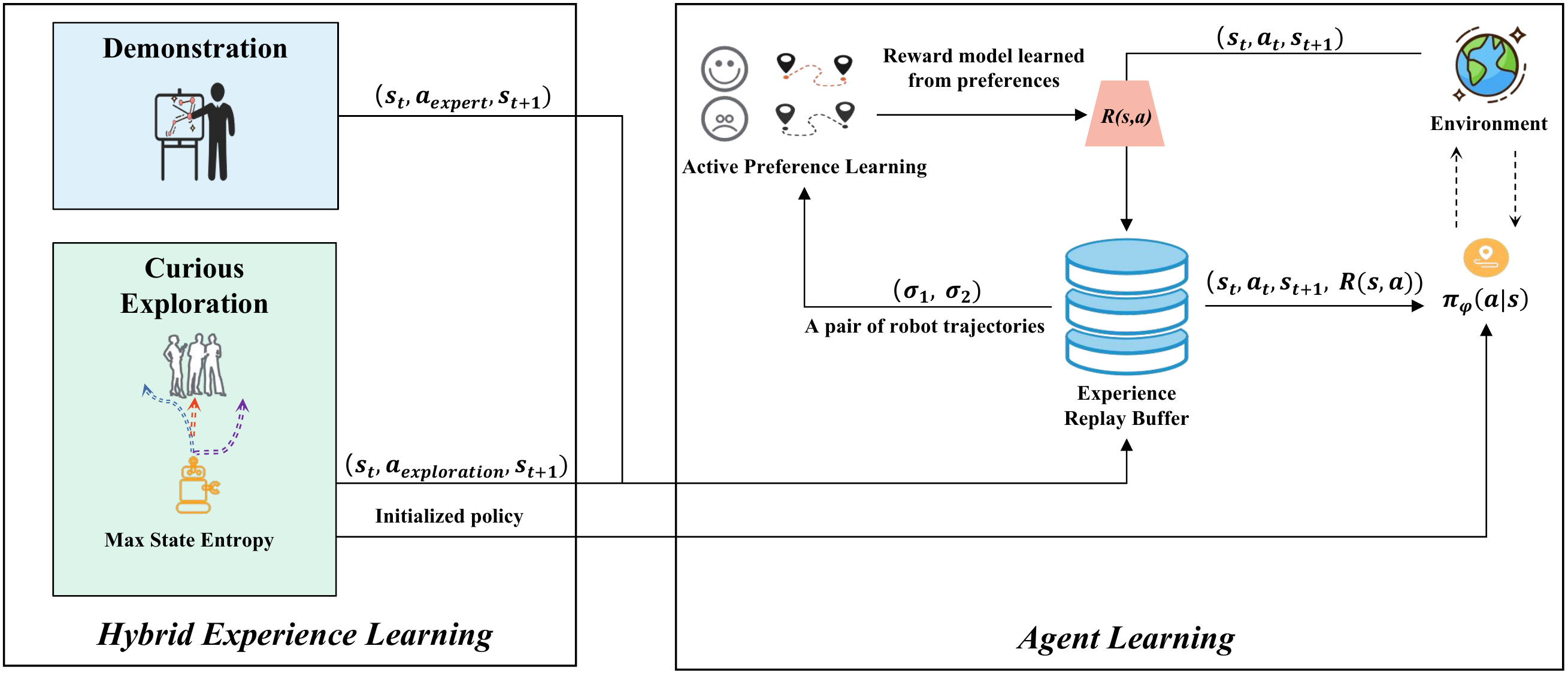}
\vspace{-10pt}
\caption{Framework of the proposed FAPL method. The FAPL is composed of two parts: 1) Hybrid Experience Learning (Section \ref{HEL}) and 2) Agent Learning (Section \ref{ARL}, \ref{OPL}). The robot agent begins with curious exploration, where it is encouraged to take different actions and reach diverse states through a state-entropy-based primitive reward. The collected exploration experiences are stored in a replay buffer along with the expert experiences from human demonstrations. Next, human teachers express preferences for a pair of robot navigation trajectories in the buffer, based on which the reward model is learned. Then, all samples are updated with a new reward value from each new generation of the reward model. At the end, the robot agent utilizes the updated samples to optimize a policy that gains maximum return from the model distilled from human preferences through off-policy RL.}
\vspace{-18pt}
\label{fig:framwork}
\end{figure*}

In this paper, we present a \textbf{F}eedback-efficient \textbf{A}ctive \textbf{P}reference \textbf{L}earning approach (FAPL) for socially aware robot navigation. The framework and procedure are illustrated in Fig. \ref{fig:framwork}. Our main idea is to efficiently distill a reward model, one having human intelligence and comfort embedded, via active preference learning. Then, the learned reward model imitates a human teacher in the subsequent RL process to guide the robot agent to explore the latent space of social compliance. To reduce the heavy human workload typically required in active preference learning, we introduce hybrid experience learning, which consists of curious exploration based on \cite{eysenbach2018diversity,liu2021behavior,lee2021pebble} and expert demonstration. Diverse exploration experiences provide a good state coverage while expert experiences guarantee positive benchmarks, enabling human teachers to give instructional feedback efficiently and accelerating the learning process.

To evaluate the performance of FAPL, we conducted extensive experiments to compare it with other four state-of-the-art baseline methods, \cite{van2011reciprocal,chen2017socially,chen2019crowd,chen2020relational}, and with one ablation model in both simulation and real-world scenarios for quantitative measurement and qualitative analysis. 


Our main contributions are: 1) To the best of our knowledge, this paper is the first to introduce active preference learning for socially aware navigation, such a purely data-driven method can tailor robot behaviors to human expectations and comfort, without the massive reward or feature engineering required in previous research; 2) The hybrid experience learning we introduce improves the efficiency of human feedback and samples; and 3) Our experiments show that FAPL can lead to more desirable and natural robot navigation behaviors.


\vspace{-5pt}
\section{Background}

\noindent\textbf{Related Work.} In optimizing robot trajectories for human comfort, existing RL-based methods \cite{chen2017socially,chen2019crowd,chen2020relational} rely on a handcrafted reward function as shown in Eq. \ref{eq1}, which penalizes collisions and uncomfortable distances. 
\begin{equation}
R_{t}=\left\{\begin{array}{ll}
-0.25 & \text {if collision} \\
-0.1+\frac{d_{t}}{2} & \text {else if distance from human}\ d_{t}<0.2 \\
1 & \text {else if reaching the goal} \\
0 & \text {otherwise}.
\end{array}\right.
\label{eq1}
\end{equation}

However, distance is not the only consideration in robot movement that influences human comfort \cite{rios2015proxemics,kruse2013human}. As such, a handcrafted reward function is unlikely to lead to robot behaviors that satisfy the broad and non-quantifiable human expectations of robots in socially compliant navigation. Handcrafted reward functions are also conducive to reward exploitation, where robots learn undesired and unnatural but highly-rewarded behaviors that impair human comfort, e.g., frequently turning 180 degrees to avoid collisions or invasive proximity. 

Another mainstream approach concerns IRL-based methods \cite{kretzschmar2016socially,okal2016learning,kim2016socially}, which aim to learn a policy or reward function through learning from human demonstration. Such human-in-loop learning approaches introduce expert intelligence to guide the robot agent, avoiding the reward engineering and exploitation characteristic of RL, and introducing naturalness to robot trajectories. Nevertheless, as the expert demonstrations constitute the only sample resources, it is very expensive to obtain sufficient and accurate samples so as to cover all latent aspects of social compliance. For instance, learning from demonstrations that lack negative data, e.g., collisions, may result in a policy that only values comfort without safety \cite{tsai2020generative}. Moreover, extensive feature engineering is required to achieve reasonable performance \cite{kretzschmar2016socially}.

Active preference or feedback learning \cite{christiano2017deep,jamieson2020active}, which relies on human feedback rather than demonstrations to provide corrective and adaptable instructions to guide the learning process, can serve as a good alternative for addressing the above-described challenges. Such interactive reinforcement learning can tailor robot behaviors to human preference naturally, without reward engineering and exploitation. In addition, it does not require and therefore is not limited by expensive human demonstrations. However, achieving good performance with active preference learning does require sufficiently frequent human feedback. Thus, it is essential to improve feedback efficiency so as to reduce the requisite human workload.

Our approach, feedback-efficient active preference learning or FAPL, can encode human comfort and expectation into a reward model and then into robot policy. To improve feedback efficiency, we introduce hybrid experience learning, which enables a human teacher to provide more critical feedback in the process of active reward learning. In contrast to existing RL methods, FAPL distills the reward model from human preferences rather than having it handcrafted, covers social compliance more intuitively and comprehensively, and avoids reward exploitation. Also, distinct from IRL methods, FAPL introduces expert intelligence without suffering from expensive, noisy, and scarce human demonstrations, nor from extensive feature engineering.


\noindent\textbf{Problem Formulation.} We follow the formulation in previous work \cite{chen2017socially,chen2019crowd,chen2020relational}. The task of navigating with humans through a spatial HRI environment is formulated as a partially observable Markov decision process (POMDP) problem in RL. Consider the movement space of all agents as a two-dimensional Euclidean space. Let $h^n_t$ present the state of $n^{th}$ human observable by the robot at time $t$. Each $h^n_t$ consists of the position: $(p^n_x, p^n_y)$ and angle: $\beta_h$ of $n^{th}$ human agent. Different from previous research, the human speed and radius are considered unobservable for better sim-to-real transfer, as these are expensive to be perceived precisely. Let $w_t$ present the the robot's state, which includes the velocity: $(v_x, v_y)$, maximum and minimum speed: $(v_{max}, v_{min})$, position: $(p_x, p_y)$, radius: $\rho$, angle: $\beta_r$ of the robot and the position of navigation goal: $(g_x,g_y)$. Then the joint state of the environment can be defined as $s^{jnt}_t$ = $[w_t,h^1_t, h^2_t, …, h^n_t]$. In each time step, the robot agent starts at an initial joint state $s^{jnt}_t$, where it takes the action $a_t$ generated by its policy $\pi (a_t|s_t)$, and then reaches the next joint state $s^{jnt}_{t+1}$ via a random unknown transition function $P(\cdot|s_t,a_t)$ from the environment and gets a corresponding reward $r_t$. The optimal policy: $\pi^{*}: \mathbf{s}_{t}^{jnt} \mapsto \mathbf{a}_{t}$, is the policy which can receive maximum expected reward return  $\mathbb{E}[\sum_{k=0}^{\infty} \gamma^{k} r_{t+k}]$, where $\gamma$ is a discount factor.

\noindent\textbf{Soft Actor-Critic Training Framework.} In this work, Soft Actor-Critic (SAC) \cite{haarnoja2018soft} will be utilized as the training framework. There are two alternating parts for parameter iteration in SAC: soft policy evaluation and soft policy improvement. In soft policy evaluation, a Q-function with parameters $\theta$, is updated to minimize Bellman-residual based objective (see Eqs. (\ref{eq2}), (\ref{eq3})), where $\mathcal{B}$ presents the replay buffer, $\gamma$ is a discount parameter, $\bar{\theta}$ and $\alpha$ are the delayed and temperature parameters respectively.
\begin{equation}
\begin{split}
\hspace{-15pt} \mathcal{L}_{Q}(\theta) &= \mathbb{E}_{\left(\mathbf{s}_{t}, \mathbf{a}_{t}\right) \sim \mathcal{B}}\bigg[\frac{1}{2}\Big(Q_{\theta}\left(\mathbf{s}_{t}, \mathbf{a}_{t}\right)- \\ 
& \left(r\left(\mathbf{s}_{t}, \mathbf{a}_{t}\right) + \gamma \mathbb{E}_{\mathbf{s}_{t+1} \sim p}\left[V_{\bar{\theta}}\left(\mathbf{s}_{t+1}\right)\right]\right)\Big)^{2}\bigg]
\end{split}
\label{eq2}
\end{equation}
\noindent where, 
\begin{equation}
\bar{V}\left(\mathbf{s}_{t}\right)=\mathbb{E}_{\mathbf{a}_{t} \sim \pi_{\phi}}\left[Q_{\bar{\theta}}\left(\mathbf{s}_{t}, \mathbf{a}_{t}\right)-\alpha\log \pi_{\phi}\left(\mathbf{a}_{t} \mid \mathbf{s}_{t}\right)\right].
\label{eq3}
\end{equation}
In soft policy improvement, the policy $\pi_{\phi}$ is optimized to minimize the following objective:
\begin{equation}
\mathcal {L}_{\pi}(\phi)=\mathbb{E}_{\mathbf{s}_{t} \sim \mathcal{B}}\left[\mathbb{E}_{\mathbf{a}_{t} \sim \pi_{\phi}}\left[\alpha \log \left(\pi_{\phi}\left(\mathbf{a}_{t} \mid \mathbf{s}_{t}\right)\right)-Q_{\theta}\left(\mathbf{s}_{t}, \mathbf{a}_{t}\right)\right]\right].
\label{eq4}
\end{equation}

\vspace{-4pt}
\section{Approach}
\vspace{-4pt}
\subsection{Overview}
In this section we introduce FAPL: Feedback-efficient Active Preference Learning. A social-compliance-embedded reward function $\hat{R}_{\mu}$, a Q-function $Q_{\theta}$  and a policy $\pi_{\phi}$ will be updated by following steps:
\begin{itemize}

	\item \textbf{Expert Demonstration:} Initially, we let human experts provide demonstrations of the socially aware navigation task and store them into an initialized experience replay buffer $EB$ as expert experiences, ($s_t$, $a_{expert}$, $s_{t+1}$). (Section \ref{HEL})
	
	\item \textbf{Curious Exploration:} We train a pre-policy  $\pi_{\phi}$ by maximizing a state-entropy-based reward to conduct curious exploration and collect diverse samples, which are stored in the buffer $EB$ as exploration experiences ($s_t, a_{exploration}, s_{t+1}$). Meanwhile, a temporary buffer $B$ is set to store incoherent samples generated during the early training process as temporary experiences ($s_t, a_{temporary}, s_{t+1}$). (Section \ref{HEL})

	\item \textbf{Active Reward Learning:} The reward model $\hat{R}_{\mu}$, encoded with human comfort and intelligence, is distilled from human feedback via active preference learning. Then all collected samples in the $EB$ and $B$ are updated with a new reward value by $\hat{R}_{\mu}$. (Section \ref{ARL})

	\item \textbf{Off-policy Learning:} The Q-function $Q_{\theta}$ and policy $\pi_{\phi}$ will be optimized using all updated samples via SAC to gain maximum return from the distilled reward model $\hat{R}_{\mu}$. (Section \ref{OPL})

\end{itemize}

\vspace{-5pt}
\subsection{Hybrid Experience Learning}
\label{HEL}
At the start of traditional active preference learning, the initialized agent follows a random policy to interact with the environment and obtain samples for a human to judge. However, such a policy covers the state space badly and leads to incoherent behaviors, making it hard for the human teacher to provide meaningful feedback \cite{liu2021behavior}. Thus, a lot of samples and human efforts are required to make initial progress. To address this challenge, we introduce hybrid experience learning module, which consists of curious exploration and expert demonstration. 

In curious exploration, we inspire the robot agent to explore the state space of socially aware navigation thoroughly to provide a better state coverage by using a $k$-NN-based state entropy estimator, adapted from \cite{singh2003nearest,liu2021behavior,lee2021pebble}, as the incipient reward function. The state entropy estimator $H_{\text {state }}(s)$, evaluates the sparsity and randomness of the state distribution by measuring the space distance between each state and its $K^{th}$ nearest neighbors as:
\begin{equation}
H_{\text {state }}(s):=\sum_{t=1}^{n} \log \left(p+\frac{1}{k} \sum_{s_{t}^{(k)} \in \mathrm{N}_{k}\left(s_{t}\right)}\left\|s_{t}-s_{t}^{(k)}\right\|\right).
\label{eq5}
\end{equation}
\noindent where $p$ is a parameter for numerical stability (usually be fixed to 1), and $s_t^{(k)}$ are the $k$ neighbors of $s_t$ in state space.

Correspondingly, the incipient exploration reward and objective function can be defined as:
\begin{equation}
R_{\mathrm{exploration}}\left(\mathbf{s}_{t}\right)=\log \left(p+\frac{1}{k} \sum_{s_{t}^{(k)} \in \mathrm{N}_{k}\left(s_{t}\right)}\left\|s_{t}-s_{t}^{(k)}\right\|\right),
\label{eq6}
\end{equation}
\begin{equation}
\phi_{\pi_{\phi}}^{\star}=\underset{\phi}{\operatorname{argmax}} \sum_{t=1}^{n} R_{\mathrm{exploration}}\left(\mathbf{s}_{t}\right).
\label{eq7}
\end{equation}

Maximizing the objective function in Eq. (\ref{eq7}), we train an initial policy $\pi_{\phi}$, by which the robot agent can explore a broader range of states. However, the socially aware navigation task not only requires the robot to reach the final goal without collisions, but values more on human comfort, which is hidden in the state space and is, thus, quite difficult to cover by the robot itself. As a result, a lot of iteration rounds and therefore a high volume of human feedback is needed to optimize a desired policy. To accelerate learning process and further improve feedback efficiency, we add human demonstrations, e.g., controlling the robot with a keyboard to demonstrate the navigation task, to introduce expert experiences to cover the hidden aspects of social compliance and provide naturalness to robot movements. The human teachers can regard expert demonstrations as benchmarks to show preferences, offering critical feedback more easily.

Meanwhile, it must be mentioned that human demonstrations are always accompanied with noise \cite{kretzschmar2016socially}. To avoid the influence of inaccurate demonstrations, we only store a triple, $(s_t, a_{expert}, s_{t+1})$, without a reward value in the replay buffer $EB$, which is different from traditional IRL. Then the human teacher can express preference on good trajectories from demonstration or dislike on inaccurate ones to add reward labels on these demonstration samples. Namely, the expert experiences are not regarded as oracles, they will be judged by human teachers like the exploration experiences, under which circumstance, we can make advantages of good demonstrations without being affected by noisy ones. The full process of hybrid experience learning is concluded in Algorithm \ref{alg:1}.


\begin{algorithm}[htb]  
	\caption{Hybrid Experience Learning.}
	\begin{algorithmic}[1]  
		\State {Initialize a critic $Q_{\theta}$ and a policy $\pi_{\phi}$}
		\State {Initialize two replay buffers ${B} \leftarrow \emptyset $ and ${EB} \leftarrow \emptyset$}
		\While {Expert Demonstration}
		\State  Store expert experiences 
		\State  $E B \leftarrow E B \cup \{(s_t,a_{\text{expert}},s_{t+1})\}$
		\EndWhile
		\While {Curious Exploration}
		\State Given $t_{\max}$ and $t_{memory}$
		\State $t \leftarrow 1$
		\While {$t < t_{\max}$}
		\State Take $a_t \sim \pi_{\phi}(a_t | s_t)$ in $s_t$ and reach $s_{t+1}$
		\State Compute reward $r_t \leftarrow R_{\text{exploration}}^{s_t}$ in (\ref{eq6})
		\If{$t<t_{memory}$}
		\State Store temporary experiences
		\State $B \leftarrow B \cup \{(s_t,a_{\text{temporary}},s_{t+1},r_t)\}$
		\Else
		\State Store exploration experiences 
		\State $E B \leftarrow E B \cup \{(s_t,a_{\text{exploration}},s_{t+1})\}$
		\EndIf
		\State Sample minibatch transitions $\sim B $
		\State Optimize $\theta$ , $\phi$ by following $\mathcal{L}_Q(\theta)$ in (\ref{eq2})
		\State and $\mathcal{L}_{\pi}(\phi)$ in (\ref{eq4})
		\State t=t+1
		\EndWhile
		\EndWhile
		\State \Return $B,EB,\pi_{\phi}$
	\end{algorithmic}  
	\label{alg:1}
\end{algorithm}  

\vspace{-2pt}
\subsection{Active Reward Learning}
\label{ARL}
The objective of active reward learning is to learn a reward function  $\hat{R}_{\mu}$, which is a neural network with parameters ${\mu}$, to encode human expectation and preference of how a socially compliant robot should act in spatial HRI. It has been shown that people feel much easier to make relative judgements than direct rating \cite{wilde2020improving}. Therefore, instead of asking human teacher to give a rate value for a single set of robot trajectories, we provide two segments for human to express preferences at a time, e.g., which segment is better or worse. We follow the framework of \cite{christiano2017deep,liu2021behavior}, to optimize our reward model $\hat{R}_{\mu}$. Robot trajectories stored in the experience replay buffer $EB$ will be divided into several segments, each segment $\sigma$ contains a sequence of states and actions in robot movements: $\sigma= ({s_t,a_t,...,s_{t+n},a_{t+n}}) $. In each feedback step, the agent will query human preference $\Upsilon$, which is one of (1,0), (0,1), (0.5,0.5), on two segments $\sigma_1$,$\sigma_2$. The judgment with two segments will be stored in a database $D$ as $(\sigma_1, \sigma_2,\Upsilon)$.

Then, a preference predictor $\mathcal{P}_\mu$ is built to train the reward model $\hat{R}_{\mu}$, where $\sigma_1\succ\sigma_2$ means that $\sigma_1$ will be preferred:
\begin{equation}
\mathcal{P}_\mu\left[\sigma_{1} \succ \sigma_{2}\right]=\frac{\exp \left(\sum_{s_t,a_t \in \sigma_{1}} \hat{R}_\mu(s_t,a_t)\right) }{\exp \left(\sum_{s_t,a_t \in \sigma_{1} \cup \sigma_{2}} \hat{R}_\mu(s_t,a_t)\right)}.
\label{eq8}
\end{equation}

\noindent The assumption behind is that the probability of a human teacher’s preference on one segment $\sigma_i$ in a pair of segments $(\sigma_i, \sigma_j)$ depends exponentially on the accumulated reward value of $\sigma_i$ gained from $\hat{R}_{\mu}$ over the whole reward value of both segments.

Based on the preference predictor $\mathcal{P}_\mu$, we optimize the reward model $\hat{R}_{\mu}$ by minimizing the cross-entropy loss function $\mathcal{L}\left(\hat{R}_{\mu}\right)$, which evaluates the difference between the prediction of $\hat{R}_{\mu}$ and ground-truth human preference:
\begin{equation}
\begin{split}
\mathcal{L}\left(\hat{R}_{\mu}\right)=-\sum_{\left(\sigma_{1}, \sigma_{2}, \mu\right) \in D} &\Upsilon(1) \log \mathcal{P}_{\mu}\left[\sigma_{1} \succ \sigma_{2}\right]+ \\ &\Upsilon(2) \log \mathcal{P}_{\mu}\left[\sigma_{2} \succ \sigma_{1}\right].
\end{split}
\label{eq9}
\end{equation}

In addition, we adapt the uncertainty-based query selection method proposed by \cite{yang2018benchmark} to determine which pair of segments $(\sigma_i, \sigma_j)$ of robot behaviors stored in $EB$ are selected to query the human teacher for preference $\Upsilon$ each time. Such an informative query selection can pick up behaviors with maximum entropy, which are typically "uncertain samples on the decision boundary", leading to a significant decrease of uncertainty in unlabelled behaviors and informative feedback from humans.

\subsection{Off-policy learning}
\label{OPL}
To further improve sample efficiency, we will adapt an off-policy RL framework, SAC, for following training. However, compared with on-policy RL frameworks, it is less stable when utilizing the reward model $\hat{R}_{\mu}$ gained from active learning, for such a reward function is always updated and thus non-static during training process, under which circumstance off-policy RL will reuse all samples in the replay buffer that contain inaccurate reward values provided by previous in-process and non-optimized reward models. To address this issue, all previous samples stored in both $EB$ and $B$ will be updated with a new reward label every time when a new reward model is distilled from human preference, and temporary experiences in $B$ will be transferred to $EB$. Then the Q-function $Q_{\theta}$ and pre-trained policy $\pi_{\phi}$ gained from curious exploration will rely on the reward model $\hat{R}_{\mu}$ combined with updated samples in $EB$ for optimization. The process of active reward learning and FAPL is presented in Algorithm \ref{alg:2} and \ref{alg:3} respectively.

\vspace{-8pt}
\begin{algorithm}[h]  
	\caption{Active Reward Learning}  		\label{alg:2}  
	\begin{algorithmic}[1]  
		\State Initialize $\hat{R}_{\mu}$ and a database $D \leftarrow \phi$
		\State Given number of feedback $M$
		\State $m \leftarrow 1$
		\State $B, EB \leftarrow$ Algorithm \ref{alg:1} 
		\While{$m<M$}
		\State Select segments $(\sigma_0,\sigma_1) \sim EB$
		\State Query human feedback $\Upsilon$
		\State Store preference $D \leftarrow D \cup \{(\sigma_0,\sigma_1,\Upsilon)\}$
		\State Sample minibatch preference $\sim D$
		\State Optimize $\mu$ by $\mathcal{L}(\hat{R}_{\mu})$ in (\ref{eq9})
		\State Update entire $EB$ and $B$ using $\hat{R}_{\mu}$
		\State $m \leftarrow m+1$
		\EndWhile
		\State Store $EB \leftarrow EB \cup B$
		\State \Return $\hat{R}_{\mu}$ and $EB$
	\end{algorithmic}  
\end{algorithm}
\vspace{-15pt}
\begin{algorithm}[h]  
	\caption{FAPL}  		
	\label{alg:3}
	\begin{algorithmic}[1]  
		\State Initialize a critic $Q_{\theta}$
		\State // Hybrid Experience Learning
		\State $\pi_{\phi} \leftarrow$ Algorithm \ref{alg:1}
		\State // Agent Learning
		\For{ each time step }
		\State //  Active Reward Learning
		\State $EB, \hat{R}_{\mu}\leftarrow$ Algorithm \ref{alg:2}
		\State Take $a_t \sim \pi_{\phi}(a_t | s_t)$ and reach $s_{t+1}$
		\State Compute reward $\hat{R}_{\mu}(s_t,a_t) \leftarrow \hat{R}_{\mu}$
		\State Store transitions 
		\State ${EB} \leftarrow EB \cup \{(s_t,a_t,s_{t+1},\hat{R}_{\mu}(s_t,a_t))\}$
		\State // Policy Optimization
		\For{each gradient step}
		\State Sample minibatch transitions $\sim EB $ 
		\State Optimize $\theta$, $\phi$ by $\mathcal{L}_Q(\theta)$ in (\ref{eq2}) and $\mathcal{L}_{\pi}(\phi)$ in (\ref{eq4})
		\EndFor
			\EndFor
		\State \Return $\pi_{\phi}$, $Q_{\theta}$
	\end{algorithmic}  
\end{algorithm}  
\vspace{-15pt}

\subsection{Implementation Details}
\label{ID}
We pretrain the policy $\pi_{\phi}$ in curious exploration for $3,000$ episodes, the samples gained in first $1,000$ episodes are stored in $B$ and the rest is stored in $EB$. In expert demonstration part, human trainers (authors) use the keyboard to control the robot velocity along the direction of x and y axis: $(v_x,v_y)$ to provide $500$ rounds of demonstrations in simulation. Once the robot is controlled to reach the goal, we say one round of demonstration is done. The reward model in active reward learning is a $3$-layer neural network with $265$ hidden units in each layer and the activation function of Leaky ReLUs. We utilize Adam to train the reward model with an original learning rate of $4\times10^{-4}$. The segment length is set to $35$ (about $15~s$), which means each segment contains $35$ tuples of $(s_t,a_t)$. And we recruited $10$ human raters, who are students or professionals in robotics, to provide $1,500$ rounds of preferences on robot behaviors for our model and ablation model respectively (\ref{AS}). Once the human rater gives a judgement $\Upsilon$ on one pair of robot trajectory segments $\sigma_1$,$\sigma_2$, one round of preference is done.

\begin{figure}[]
\centering
\vspace{-5pt}
\includegraphics[width=\columnwidth]{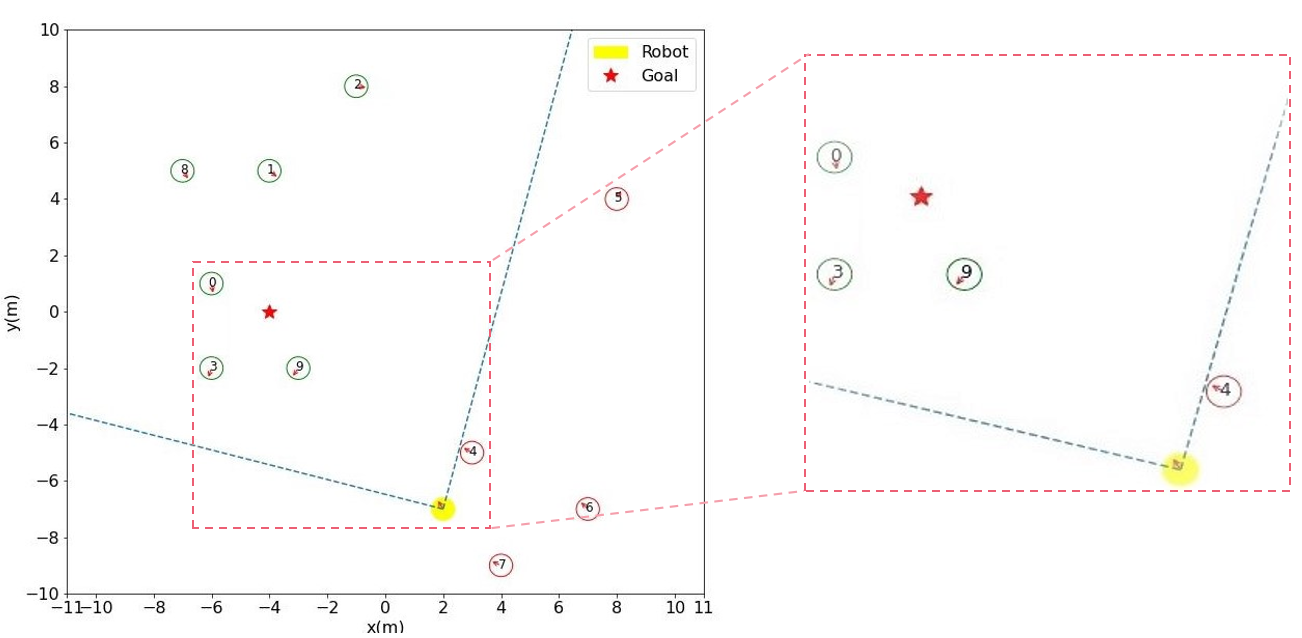}
\vspace{-20pt}
\caption{Simulation environment illustration. In a $22~m\times20~m$ two-dimensional space, the yellow circle indicates the robot. The blue dotted line illustrates the robot FoV, humans that can be detected by the robot are green circles while those out of robot view are red circles. The red star is the robot goal, and the orientation and number of each agent are presented as a red arrow and a black number respectively.}
\vspace{5pt}
\label{fig3}
\end{figure}
\vspace{-3pt}

\section{Experiments And Results}
\subsection{Simulation Experiment}
\subsubsection{Simulation Environment}
We adapted a simulation environment from \cite{chen2017socially,chen2019crowd} for training and experiments as is shown in Fig. \ref{fig3}. Each agent in the environment follows holonomic kinematics, where agents can move to any direction at any time. The agent’s action at time $t$ is the preferred velocity in x-axis and y-axis direction: $a_t=(v^x_t,v^y_t)$. Such a velocity is assumed to be immediately achievable. All human agents are controlled by ORCA \cite{van2011reciprocal}, and the parameters of their policy are generated by a Gaussian distribution to provide random pedestrian behaviors, e.g., different velocity and goals. To avoid learning the exceedingly aggressive behavior that the robot compels all humans to yield, we follow previous research \cite{chen2019crowd} to set an invisible robot setting, where the human agent is required only to do reaction, e.g., yield, to other human agents. Meanwhile, different from previous setting, we set the degree of robot field-of-view (FoV) as $90^{\circ}$ rather than $360^{\circ}$ to better mimic real-world scenarios and reduce sim-to-real domain-mismatch issues, since sorting multiple sensors on a physical robot to obtain a global FoV is quite impractical and expensive.

\subsubsection{Baselines}
We compare the proposed FAPL with other four state-of-the-art methods mentioned before: ORCA \cite{van2011reciprocal} is regarded as the baseline of model-based approaches; CADRL \cite{chen2017socially}, SARL \cite{chen2019crowd} and RGL \cite{chen2020relational} are regarded as the baselines of learning-based approaches. 

\subsubsection{Ablation Study}
\label{AS}
To evaluate the feedback efficiency of FAPL, we also implement another ablation model APL, which removes the hybrid experience learning module in FAPL, as the baseline of active preference learning. 

\subsubsection{Training Details}
We utilize the same reward function in \ref{eq1} for CADRL, SARL and RGL. The architectures of all networks stay the same in each experiment. All baseline networks are trained by following the implementing details in original papers for $1\times10^4$ episodes. We train APL and FAPL with $1,500$ pieces of human feedback respectively (\ref{ID}) for $1\times10^4$ episodes using a learning rate of $2\times10^{-4}$ with same parameters.  

\subsubsection{Evaluation}
We compare the learning curves of all learning-based methods in terms of success rate and discomfort frequency, which refers to the percentage of duration where the robot is too close with a human pedestrian. Three experiment setting: $5$, $10$, $15$ motional human agents are involved respectively, are set to evaluate the performance of all methods. In each setting, $500$ unseen  testing scenarios are tested with five indicators recorded: success rate, time-out rate, collision rate, discomfort frequency and time to reach the goal. The time-out threshold is set as $30$, $35$ and $40$ seconds in each setting separately, and the margin of discomfort frequency is set as $0.3~m$ based upon \cite{rios2015proxemics}. 

\subsubsection{Results}

\begin{figure}[t]
\centering
\subfloat[Success Rate]{\includegraphics[width=0.5\columnwidth]{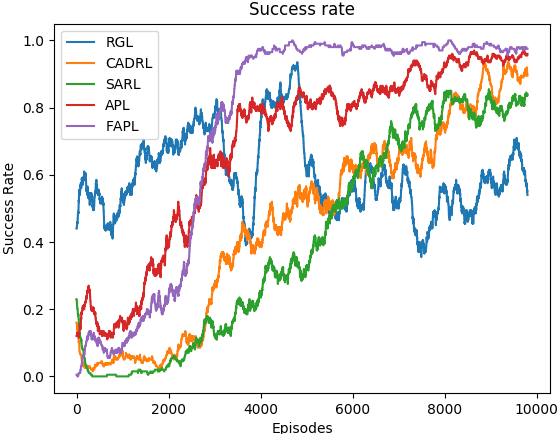}}
\subfloat[Discomfort Frequency]{\includegraphics[width=0.5\columnwidth]{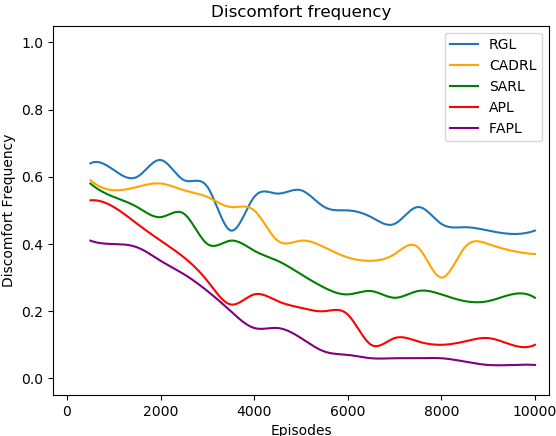}}
\caption{Learning curves of CADRL, RGL, SARL, APL and FAPL as measured, on (a) success rate (each episode) and (b) discomfort frequency (the mean across each five hundred episodes) under the setting of 10 humans.}
\label{fig4}
\end{figure}

Fig. \ref{fig4} shows the learning curves of all learning-based methods in terms of success rate and discomfort frequency. FAPL (purple) can improve the success rate and reduce discomfort frequency continuously, and achieve the best performance in the end of training, outperforming all baselines significantly. Much as RGL (blue) achieves a higher success rate than ours in first $3,000$ episodes, it is very unstable and finally only reaches a success rate of about $0.6$ in the end. Compared with APL (red), FAPL has a lower success rate and higher discomfort frequency in the beginning, for its initial policy comes from curious exploration where the robot is only encouraged to visit different states. However, FAPL surpasses APL after around $3,000$ episodes and achieves the best performance much more quickly, which shows that FAPL can converge significantly faster with the same number of human feedback, quantitatively proving the benefits of the hybrid experience learning module.

\begin{figure*}[htb]
\centering
\vspace{-5pt}
\includegraphics[width=16cm]{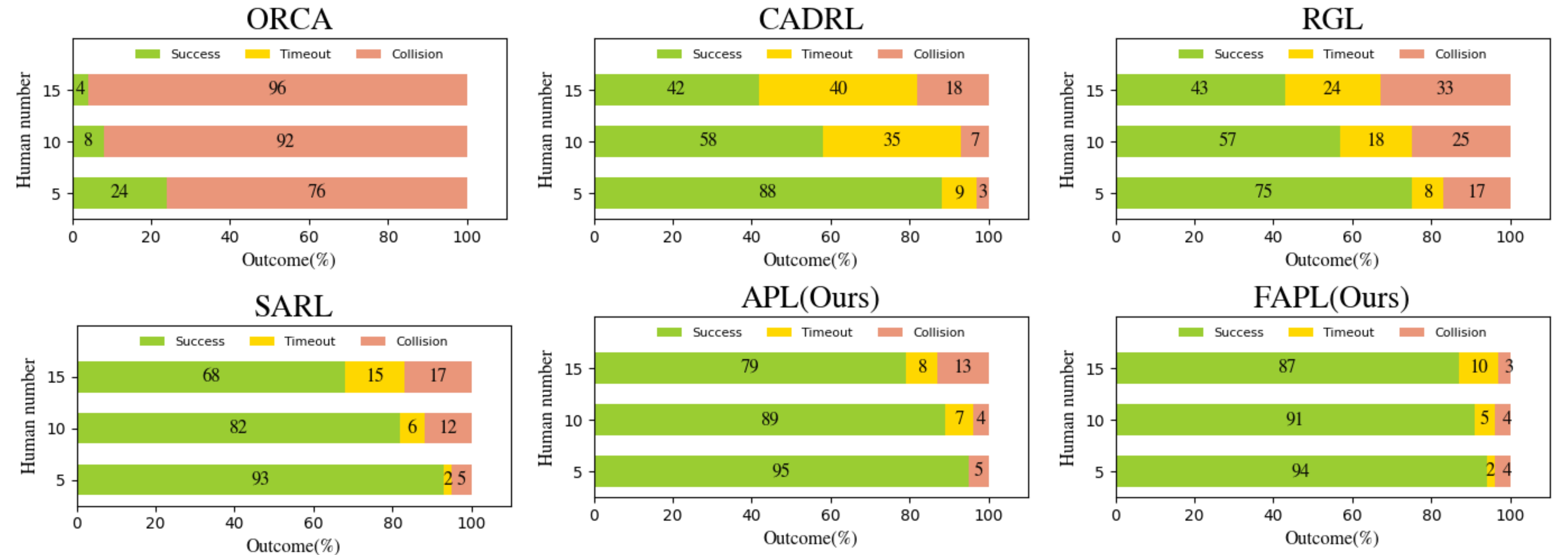}
\vspace{-8pt}
\caption{Outcome: rates of success, collision and time-out under three experimental settings with five, ten and fifteen human agents involved respectively. }
\vspace{-14pt}
\label{fig5}
\end{figure*}

Fig. \ref{fig5} presents the percentage of success, time-out, and collision rates of each model. ORCA performs quite badly as expected, for its model is based on the assumption that all agents are always reciprocal, violating the invisible robot setting. RGL also performs badly since it is a model-based RL built based upon global robot FoV and precise perception of human velocity which are hard to obtain in our setting. Meanwhile, the learning based-baselines are competitive to our FAPL and APL in the setting of $5$ humans. However, with the increase of the complexity of environment, i.e., the number of humans, our methods perform better and better than the baselines. This shows that APL and FAPL are more efficient and robust in complex task scenarios. And FAPL outperforms APL apparently in the most difficult scenario involving $15$ humans.

\begin{table}[t]
\caption{Outcome: time (second) and discomfort frequency (DisFq.).\label{tab:table1}}
\centering
\begin{scriptsize}
\begin{tabular}{cccccccc}
\hline & \multicolumn{3}{c}{ Time } & & \multicolumn{3}{c}{ DisFq. } \\
\cline { 2 - 4 } \cline { 6 - 8 } Methods & \multicolumn{3}{c}{ Human Number } & & \multicolumn{3}{c}{ Human Number } \\
& 5 & 10 & 15 & & 5 & 10 & 15 \\
\hline 
CADRL & $29.4$ & $34.7$ & $41.3$ & & $0.203$ & $0.258$ & $0.435$ \\
SARL & $27.3$ & $32.5$ & $38.7$ & & $0.137$ & $0.172$ & $0.251$ \\
RGL & $\mathbf{22.9}$ & $\mathbf{27.4}$ & ${32.7}$ & & $0.288$ & $0.354$ & $0.474$ \\
APL(Ours) & ${24.2}$ & $28.7$ & $\mathbf{31.9}$ & & $0.029$ & $0.043$ & $0.061$ \\
FAPL(Ours) & $25.3$ & $30.1$ & $34.1$ & & $\mathbf{0.018}$ & $\mathbf{0.025}$ & $\mathbf{0.048}$\\
\hline
\end{tabular}
\end{scriptsize}
\label{table_1}
\end{table}

Table \ref{table_1} demonstrates the discomfort frequency and navigation time of each model except ORCA due to its extremely high collision rate. RGL enjoys the  best overall performance in navigation time, however, it has the highest discomfort frequency, which means, to achieve a shorter time, the robot takes a lot of aggressive action, hurting human comfort badly. FAPL obtains the lowest discomfort frequency with shorter time than CADRL and SARL, and is competitive to RGL and APL in terms of navigation time. Combining the indicators of navigation time and discomfort frequency shows that our methods lead to robot behaviors that respect human comfort distance more than all baselines and ablation model, without sacrificing too much time due to lazy behaviors, e.g., waiting still for all pedestrians to pass or excessive detouring.

\subsection{Real-world Experiment}

Since the social compliance is non-quantifiable and broader than the comfortable distance, it is unsatisfied and insufficient to evaluate it only via the indicator of discomfort frequency in simulation experiments. To intuitively and further evaluate the social compliance of learned robot trajectories from our methods, we recruited human participants to conduct real-world experiments, collecting their feedback from experiences of walking with a robot controlled by different models as another indicator. This 
real-world experiment has been reviewed by Institutional Review Board (IRB) of BUCT.

\subsubsection{Robot and Environment Setup}
\begin{figure}[t]
\centering
\includegraphics[width=0.95\columnwidth]{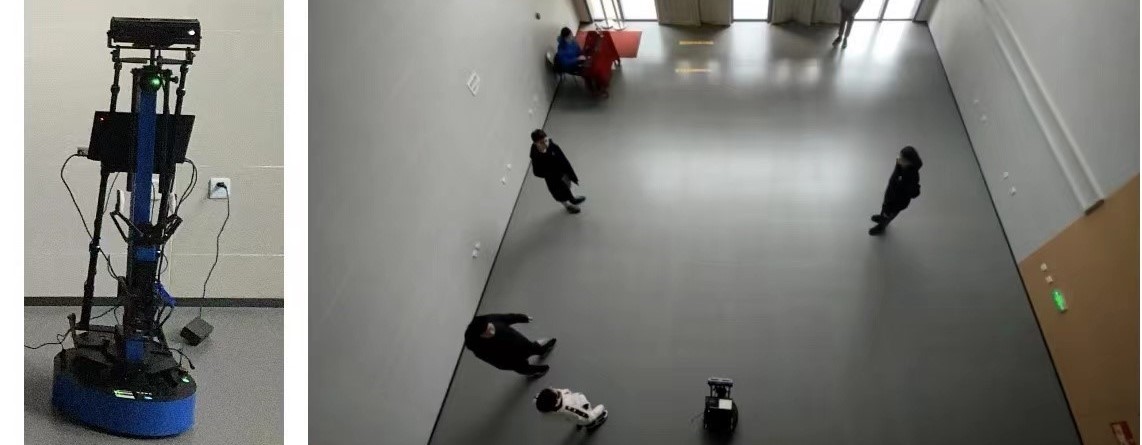}
\vspace{-5pt}
\caption{Illustration of the robot platform (left) and environment (right) employed for the real-world experiment.}
\vspace{5pt}
\label{fig6}
\end{figure}

\begin{figure}[t]
\centering
\includegraphics[width=0.95\columnwidth]{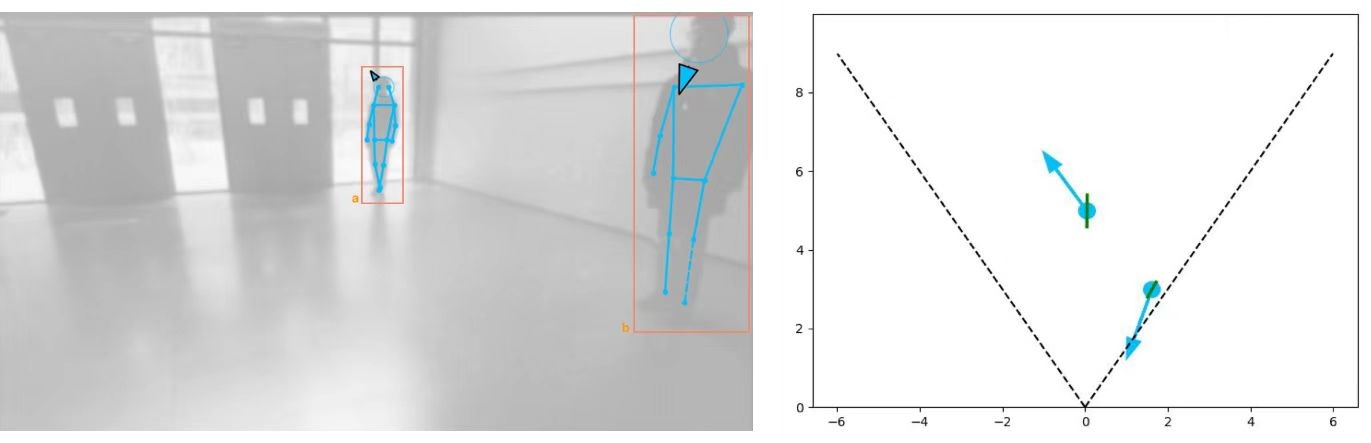}
\vspace{-5pt}
\caption{Illustration of perception algorithms for human tracking (left) and localization (right), where the blue circle and arrow present the position and orientation of a human respectively.}
\vspace{5pt}
\label{fig7}
\end{figure}

We utilized an Enlighten mobile robot \cite{6-robot} (Fig. \ref{fig6}, left) as the platform for real-world experiments. A Kinect v2 camera with an approximately field-of-view of $84.1^{\circ}$ was set on the top of the robot to capture human positions. YOLOv5 \cite{YOLOv5} combined with DeepSORT \cite{DeepSort}, and Monoloco Library \cite{Monoloco_library,bertoni2021perceiving} were used for robot-centered human tracking and localization (Fig. \ref{fig7}). To provide sufficient real-time computing power for the perception algorithms and RL controllers, we connected the robot to a host with a RTX 3090 GPU via ROS. To avoid potential damage to the robot hardware, we set up an action restriction: the robot is enforced to stop when its action contains a turning degree greater than $90^{\circ}$. The experimental environment was an approximate $23~m\times 16~m$ controlled open space as shown in Fig. \ref{fig6}, right. 

\subsubsection{Baselines}
The ablation model APL and SARL, the best performing baseline in terms of discomfort frequency in simulation, were selected as the baselines.

\subsubsection{Experimental Design} 
We recruited $10$ participants ($2$ females and $8$ males), all aged over $18$ years old ($\mu$~=~$21.9$; $\sigma$~=~$1.92$), and divided them into $2$ groups. During each experiment, the robot and participants as pedestrians were required to perform point-to-point navigation tasks in the same environment. The human pedestrians were encouraged to behave naturally without any action restrictions during tasks. We set up three scenarios, where the robot and human pedestrians were assigned with different starting positions and goals. We asked each group to spatially interact with the robot controlled by three models: SARL, APL, and FAPL respectively in the three different scenarios, resulting total $18$ tests. The order of three models was randomly set in each scenario for each group, and all participants were unaware of which model was active during each test. 

\subsubsection{Evaluation}
Once each test was done, the participants were asked to fill a questionnaire to rate robot behaviors of three models in terms of comfort and naturalness using Likert Scale \cite{joshi2015likert} with 1 being strongly disagree and 5 being strongly agree, based on their experiences of spatial interaction with the robot. The participant's responses to the questionnaire were used for quantitative measurement. We also conducted a semi-structured interview to collect oral responses of participant's experience in experiments for qualitative analysis. 

\subsubsection{Quantitative Measurement}
\begin{figure}[h]
\centering
\includegraphics[width=\columnwidth]{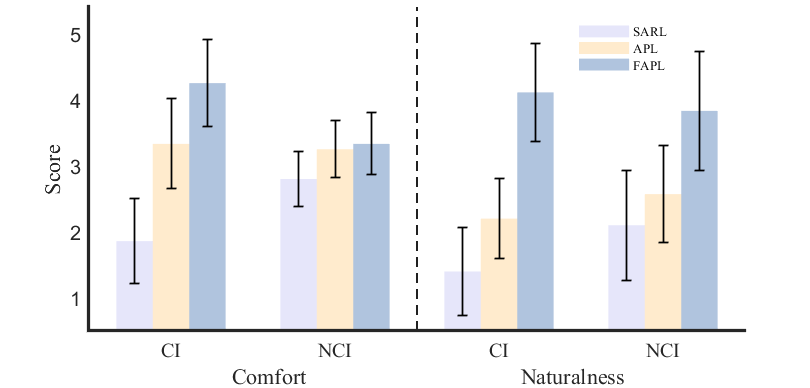}
\vspace{-20pt}
\caption{Summary of participant responses to evaluate robot navigation behaviors in terms of comfort and naturalness.}
\vspace{5pt}
\label{fig8}
\end{figure}

During each test, participants were divided into two categories: close interaction participant (CI) and non-close interaction participant (NCI), to collect fair questionnaire responses, where CI refers to pedestrians who approach the robot within $0.6~m$ for more than $2$ seconds during the test while NCI refers to those who do not. The results are summarized in Fig. \ref{fig8}. The benefits of robot behaviors of FAPL in terms of comfort and naturalness can be obviously seen from the responses of CI. The reason why FAPL does not outperform the other two much from the responses of NCI for comfort is that NCI did not have enough direct experiences to judge, and therefore, they tended to give all three models a middle score. And the reason why responses for naturalness of CI and NCI are similar is that the robot unnatural behaviors, e.g., suddenly stopping, were accompanied with noises that attracted NCI's attention.

\subsubsection{Qualitative Analysis}
After each test, all participants were asked an open-ended interview question that: "Could you share your experiences and explain reasons behind?". The insights thematically coded from participants' responses are presented along with supporting quotes as follows.

\noindent\textbf{Long-sighted Decision.} One reason why the robot behavior in FAPL is preferred by pedestrians is that it takes more long-sighted decisions, avoiding potential risky situations where the robot is hard to maintain socially compliant interaction with pedestrians. We believe this is because we introduce human intelligence and insights via expert demonstration and feedback in the learning process, enabling the robot to handle complex situations by valuing more on long-term benefits than short-term gains. Some notable quotes are as follows:

(P$5$) "The reason why I like that one (FAPL) is it turned right to give me enough space before I got close to it."

(P$9$) "Because the third one (FAPL) went left to detour at the start instead of getting involved in the center of the crowd directly, like the first one (SARL)."

\noindent\textbf{Behavior Naturalness.} Another reason why our FAPL is preferred is that it leads to more natural behaviors, e.g., does not stop suddenly or turn left and right frequently. We owe this to the human-imitative trajectory samples introduced by expert demonstration in hybrid experience learning, and the avoidance of reward exploitation problems by replacing handcrafted reward functions with reward models distilled from human preference. Some notable quotes are as follows:

(P$1$) "The second one (SARL) always stopped immediately, and the first (APL) also did that sometime, these made me a little nervous and annoyed."

(P$9$) "For me, that one (FAPL) moved more like a human, I mean it hardly stopped suddenly and turned frequently, which is preferable to me."

\section{Conclusion}
In this work, we present FAPL, a feedback-efficient active preference learning approach for socially aware robot navigation that distills human comfort and expectation into a reward model to guide the robot to explore the latent space of social compliance. We demonstrate via both simulation and real-world experiments that our method outperforms existing state-of-the-art approaches, leading to more desirable and natural robot behaviors. We also show that by introducing hybrid experience learning, the efficiency of human feedback can be improved. 


\vspace{-5pt}
\section*{ACKNOWLEDGMENT}
This material is based upon work supported by the National Science Foundation under Grant No. IIS-1846221. 

\typeout{}
\bibliography{main}
\bibliographystyle{IEEEtran}
\end{document}